\documentclass[conference]{IEEEtran}

\usepackage{graphicx}
\usepackage{amsmath,amssymb}
\usepackage{siunitx}
\usepackage{upgreek}
\usepackage[noend]{algpseudocode}
\usepackage{algorithm}
\usepackage{multirow}
\usepackage{url}
\usepackage{balance}
\let\labelindent\relax
\usepackage{enumitem}
\usepackage{xcolor}
\usepackage{soul}
\usepackage{hyperref}
\usepackage{caption}
\usepackage{subcaption}
\usepackage{booktabs}
\usepackage{makecell}
\usepackage[colorinlistoftodos,prependcaption,textsize=small]{todonotes}
\usepackage{longtable}
\usepackage{lscape}
\usepackage{hvfloat}
\usepackage[table]{xcolor}

\usepackage{cite}  

\ifCLASSINFOpdf
\else
\fi

\begin{document}
\bstctlcite{IEEEexample:BSTcontrol}
%

\title{\vspace{-1ex}Learning Parametric Nitrogen Fertilizer Response Curves Using Neuro Symbolic Regression }

\author{\IEEEauthorblockN{Giorgio Morales$^{1}$ and John Sheppard$^{2}$}
\IEEEauthorblockA{$^{1}$ Aston Centre for Artificial Intelligence Research and Application, Aston University, Birmingham, UK \\
$^{2}$ Gianforte School of Computing, Montana State University, Bozeman, US}\vspace{-3.2ex}
}


\maketitle

\begin{abstract}
Accurately modeling crop response to Nitrogen (N) fertilization is a fundamental challenge in precision agriculture, as it impacts both economic returns and environmental sustainability. 
Existing approaches either rely on predefined parametric forms or opaque machine learning models, limiting their ability to interpret or discover site-specific functional relationships from data.
In this work, we propose a neuro symbolic regression (SR) approach to learn parametric N-response curves without assuming a predefined functional form. 
Our approach integrates a transformer-based Multi-Set Symbolic Skeleton Prediction strategy, enabling the discovery of shared functional structures across multiple subdomains or management zones (MZs). 
By constructing diverse input subsets and enforcing consistency across them, the method recovers robust symbolic skeletons that are subsequently fitted to observed data using a genetic algorithm.
This framework was first evaluated on synthetic one-dimensional problems to assess its robustness under varying levels of epistemic uncertainty. 
The results demonstrate the ability of the proposed SR approach to recover correct expressions even in data-scarce regimes.
In this work, we present the results of applying our method to real-world winter wheat data, learning distinct parametric N-response curves for different MZs within a field. 
The results show that the discovered expressions not only achieve lower fitting errors than traditional models such as quadratic-plateau and exponential functions, but also capture diverse functional behaviors across spatial regions.
This demonstrates the potential that neuro SR has to enable the discovery of site-specific agronomic relationships and support informed decision-making in precision agriculture. 
\end{abstract}



\section{Introduction} \label{sec:intro}

In recent years, the field of agriculture has undergone a significant transformation due to advances in technology~\cite{yieldML} and the need for more sustainable and efficient practices~\cite{hegedus}. 
In that context, precision agriculture (PA) addresses key agricultural challenges by leveraging data to optimize operations, 
reduce environmental impact, and maximize profit~\cite{PAreview}.
To accomplish these objectives, models are trained to represent relationships between input covariate factors, gathered from sensors and agricultural machinery, and outcome variables, such as crop yield~\cite{Morales_2023}. 
Subsequently, these models are used to predict and analyze how outcome variables change across different rates of spatially variable input factors. 
Thus, the purpose of these models is to provide valuable insights for making informed and optimal decisions in agricultural management~\cite{amyoptim,amysust}. 
A critical aspect in the development of these models and simulations is On-Farm Precision Experimentation (OFPE), a framework that yields site-specific data about how fields respond to various management practices~\cite{hegedus,cook2018}.

Within PA, symbolic regression (SR) has been relatively underutilized despite its potential to aid the agricultural objectives. 
SR seeks to discover mathematical equations that characterize the underlying dynamics of a system. 
These equations, which aim to be concise and interpretable, not only offer the means to make accurate predictions and informed decisions but also provide important insights into the system. 
In PA, farmers need to understand how crops respond to input factors precisely. 
Among these, nitrogen fertilizer rate (N-rate) and seeding rate receive particular attention due to their role in crop management and resource optimization~\cite{maxwell,bullock}.

SR, with its potential to be used as a scientific discovery tool, offers a promising solution to this problem. 
This work explores the application of SR to PA.
Specifically, the main objective is the discovery of accurate mathematical equations that encapsulate the dynamics between the N fertilizer and the yield response of a constructed fertilizer management zone (MZ). 
By doing so, we aim to equip farmers with valuable tools for informed decision-making and resource optimization.

We present one approach to solving
the problem of distilling parametric, site-specific N fertilizer–yield response (N-response) curves from opaque models such as deep neural networks (NNs). By focusing on MZs with consistent field characteristics, we can abstract
N-response curves within those MZs to be one-dimensional (1-D), defined over a fixed range of N-rates. 
In PA, N-response curves serve as a tool to calculate the Economic Optimum Nitrogen Rate (EONR)~\cite{bullock94}; i.e., the N rate beyond which further N application ceases to be profitable. 
Furthermore, substituting these lightweight equations for computationally expensive NNs enables highly efficient downstream multi-objective optimizations~\cite{amysust}.

However, due to the typically limited data available in agricultural settings, these curves are often affected by epistemic uncertainty in certain regions of the domain. 
Thus, Sec.~\ref{sec:1dmethod} introduces an SR methodology designed to extract mathematical expressions from 1-D curves under epistemic uncertainty. 
This methodology is evaluated on synthetic 1-D problems within an adaptive sampling (AS) framework, which progressively augments the dataset to reduce uncertainty. 
The experiments demonstrate the stability and convergence of the discovered expressions as uncertainty is reduced.

Then, Sec.~\ref{sec:Nresponse} applies the proposed methodology to a real agricultural case study. 
Given a winter wheat dataset with site-specific N-response curves generated by a pre-trained early yield prediction NN, along with a set of predefined MZs, our objective is to recover the underlying functional form associated with each MZ. 
The resulting symbolic expressions are then fitted to individual field sites to model their corresponding N-response curves. 
Experimental results show that the generated curves closely match those produced by the NN while achieving lower prediction errors compared to traditional approaches based on predefined functional forms.

\section{Background} \label{sec:background}

This section provides an overview of the SR framework that serves as the backbone of our approach, namely Symbolic Regression using Transformers, Genetic Algorithms, and Genetic Programming (SeTGAP). 
It also introduces essential concepts related to agricultural MZs and related agronomic definitions.

\subsection{Decomposable Neuro Symbolic Regression}

SeTGAP~\cite{SeTGAP}
addresses multivariate SR problems by probing trained opaque regression models to distill them into multivariate mathematical expressions that allow us to interpret the regression model’s functional form.
%
SeTGAP starts by producing univariate symbolic skeletons that estimate functional relationships for each variable with respect to the response variable.
A symbolic skeleton $\mathbf{e}(x_v)$ is an abstract representation that captures the functional form between $x_v$ and the system's response using constant placeholders that may represent numeric constants or functions of other variables.
For example, if $f(\mathbf{x}) = {5 \log{x_1}}{\,(\sin (x_2^2) + 1)} -4$, then $\mathbf{e}(x_1) = c_1{(\log{x_1})} + c_2$, and $\mathbf{e}(x_2) = c_3\sin(x_2^2) + c_4$. 

When analyzing a specific variable, SeTGAP generates multiple input–response pairs by fixing the remaining system variables to sets of randomly sampled values. 
Each resulting set enables examining the relationship between the variable of interest and the system’s response under varying contextual conditions.
The task of predicting a skeleton $\hat{\mathbf{e}}(x_v)$ that describes the shared function form of $N_S$ sets, each consisting of $n$ input–response pairs (i.e., $\hat{\mathbf{e}}(x_v) \approx \mathbf{e}(x_v)$), is known as multi-set symbolic skeleton prediction (MSSP).

The MSSP problem is tackled using a Multi-Set Transformer, pre-trained on a large corpus of symbolic skeleton expressions.
The obtained univariate skeletons are considered candidate subexpressions that are then merged incrementally into multivariate expressions that approximate the system’s
underlying equation, employing GA- and GP-based techniques while preserving the originally identified skeleton structures.

\subsection{Agricultural Management Zones}

MZs are subregions within a field with similar characteristics affecting crop yield. 
They account for spatial variability in factors such as soil properties, leading to different management needs across the field. 
By treating these areas separately, farmers can tailor practices like fertilizer application to improve productivity while minimizing environmental impact~\cite{managementNrate}.

In prior work~\cite{MZ}, 
we presented the first MZ delineation approach that accounts for within-field variability of fertilizer responsivity based on NN-generated N-response curves.
A clustering strategy leverages the shape dissimilarity of N-response curves, characterized using functional principal component analysis.
This allows for establishing MZs whose sites show comparable responsivity to varying N rates.

Existing \textit{post-hoc} interpretability methods based on counterfactual explanations~\cite{responsivity,MZ} 
have been proposed to understand the influence of covariate factors on N responsivity and MZ assignments. 
However, they do not provide a thorough understanding of the mathematical behavior of the field's N responsivity at various locations.
Moreover, to the best of our knowledge, no prior work has studied learning the mathematical expressions that describe site-specific N-response curves from data without pre-specifying their functional form.

\section{1-D Equations Under Epistemic Uncertainty}  \label{sec:1dmethod}

Consider a dataset $\mathcal{D} = \left(\mathbf{X}_{obs}, \mathbf{y}_{obs}\right)$ and a model $\hat{f}$ trained on it. 
Given that $\mathbf{X}_{test}$ defines a fixed input domain grid, the model prediction $\hat{\mathbf{y}}_{test} = \hat{f}(\mathbf{X}_{test})$ represents the estimated response across that domain.
To derive a mathematical interpretation of the function computed by $\hat{f}$ in the form of an expression, we analyze the set $\hat{\mathcal{D}} = \left(\mathbf{X}_{test}, \hat{\mathbf{y}}_{test}\right)$, which pairs the input domain with the model’s estimated response.


\subsection{Methodology}

\begin{figure}[t]
    \centering
    \includegraphics[width=.75\linewidth]{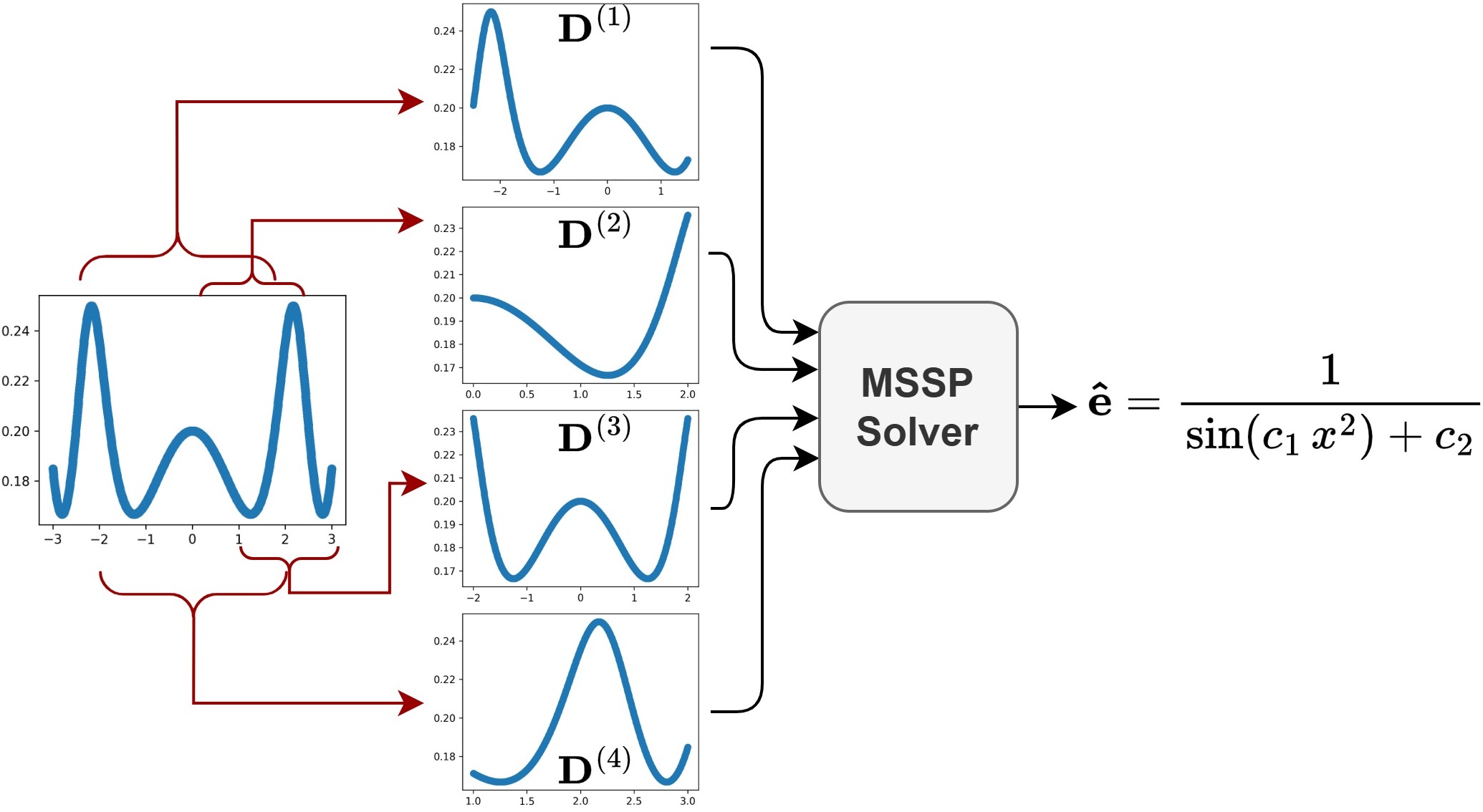}
    \vspace{-1ex}
    \caption{MSSP example for solving a 1-D problem.}
    \vspace{-2ex}
    \label{fig:curveMSSP}
\end{figure}

SeTGAP was originally designed to solve multivariate SR problems in a decomposable way.
However, the problem tackled in this work belongs to a 1-D setting due to the abstractions arising from focusing on MZs. 
In such cases, a symbolic skeleton prediction method~\cite{bertschinger_evolving_2024,SRthatscales,petersen2021deep,SymbolicGPT} can be used to infer a univariate expression from a single input-response pair, capturing the relationship between $\mathbf{X}_{test}$ and $\hat{\mathbf{y}}_{test}$. 
Alternatively, we can replicate the set $\hat{\mathcal{D}}$ $N_S$ times to form a collection of sets suitable for processing by our Multi-Set Transformer $g(\cdot)$.
Nevertheless, to leverage SeTGAP’s MSSP capabilities, we adopt a different strategy.
Consider the curve shown in Fig.~\ref{fig:curveMSSP}, generated from the function $f(x) = \frac{1}{\sin(x^2) + 5}$.
Rather than simply replicating $\hat{\mathcal{D}}$ $N_S$ times (with $N_S = 4$ in this example), we construct a collection of distinct sets $\mathbf{D} = \left\{ \mathbf{D}^{(1)}, \dots, \mathbf{D}^{(N_S)} \right\}$ by randomly sampling subsets from the original input domain.
Since each set in $\mathbf{D}$ is derived from the same original curve, they all share a common symbolic skeleton. 
The task of the MSSP solver is then to recover a skeleton $\hat{\mathbf{e}}$ that approximates this shared structure.

This strategy introduces diversity into the SR problem, helping to mitigate localized uncertainty. 
For instance, regions with sparse data or increased noise may exhibit distortions and cause traditional methods 
to overfit by generating unnecessarily complex functions. 
By sampling distinct subregions and seeking a shared symbolic structure, the MSSP framework emphasizes the recovery of the underlying functional form rather than the noise-driven anomalies of specific regions.
This approach is most effective when the input domain is broad enough to exhibit meaningful variation across subregions.
That is, in cases with limited domain coverage or low variability, alternative techniques may be more appropriate.

Algorithm~\ref{alg:1DMSSP} presents a variant of SeTGAP's univariate skeleton prediction algorithm. 
Function $\texttt{generate1DExpr}$ takes as input the datasets $\mathcal{D}$ and $\hat{\mathcal{D}}$, the Multi-Set Transformer $g$, the number of input sets $N_S$, the number of points per set $n$, and two hyperparameters, $n_B$ and $n_{\text{cand}}$, which control the number of skeleton candidates generated. 
On line~\ref{line:range}, the function \texttt{getRandomRange} randomly selects a portion of the input domain, between 80\% and 100\%, to construct the $i$-th input set. 
Function \texttt{selectAndInterpolate} is then used to extract the subset of data points in $\hat{\mathcal{D}}$ that lie within the selected input range. 
If the number of points in this region is insufficient, linear interpolation is applied to resample the data and produce exactly $n$ points.
The multiple generated sets allow us to construct the collection $\mathbf{D}$, which is processed by the model $g$ to generate $n_B$ outputs. 
This procedure is repeated $n_{\text{cand}}$ times to obtain a diverse set of candidate skeletons. 
After generating $n_{\text{cand}}$ sets of skeletons, the coefficients of each candidate are fitted to the observed data $\mathcal{D}$ using the GA-based optimizer \texttt{fitCoefficients}, which minimizes the MSE between the fitted expression and the target values. 
The final output is the fitted expression $\tilde{f}$ with the lowest error.

\setlength{\textfloatsep}{8pt}
\begin{algorithm} [!t]
\scriptsize
\renewcommand{\algorithmicrequire}{\textbf{Input:}}  
\renewcommand{\algorithmicensure}{\textbf{Output:}} 
\begin{algorithmic}[1]
    \Require $\mathcal{D} = (\mathbf{X}_{obs}, \mathbf{y}_{obs})$; domain grid $\mathbf{X}_{test}$ and estimated response $\hat{\mathbf{y}}_{test}$; Multi-\hspace*{1.6em}Set Transformer $g$; \# of input sets $N_S$; \# of candidates $n_{\text{cand}}$; beam size $n_B$   
    \Ensure Estimated expression $\tilde{f}(\mathbf{x})$ and corresponding skeleton $\tilde{\mathbf{e}}(\mathbf{x})$
    \vspace{-1.5ex}
    \Statex \makebox[0pt][l]{\hspace*{-\algorithmicindent}\rule{\dimexpr\linewidth+\algorithmicindent}{0.4pt}}
    
\Function{generate1DExpr}{$\mathbf{X}_{obs}, \mathbf{y}_{obs}, \mathbf{X}_{test}, \hat{\mathbf{y}}_{test}, g, N_s, n, n_{\text{cand}}, n_B$}
    \State $\text{genSks} \leftarrow [\,]$ 
    \For { each $i \in (1, n_{\text{cand}})$}
        \State $\mathbf{D} \leftarrow [\,]$ 
        \For { each $j \in (1, n_{\text{cand}})$}
            \State $\text{range} \leftarrow \texttt{getRandomRange}(\mathbf{X}_{test})$ \label{line:range}
            \State $\mathbf{X}^{(s)},\mathbf{y}^{(s)} \leftarrow \texttt{selectAndInterpolate} \label{line"interpolate}(\mathbf{X}_{test}, \hat{\mathbf{y}}_{test}, \text{range}, n)$
            \State $\mathbf{D}.\texttt{append}\left( (\mathbf{X}^{(s)},\mathbf{y}^{(s)}) \right)$
        \EndFor
        \State $ \text{genSks}.\texttt{append}(g(\mathbf{D}, \Theta; n_B))$
    \EndFor
    \State $\text{genSks} \leftarrow \text{removeDuplicates}(\text{genSks})$
    \State $\text{MSEvals}, \text{genExps} \leftarrow \text{zeros}(|\text{genSks}|), \text{zeros}(|\text{genSks}|)$
    \For { each $k \in (1, n_{\text{cand}})$} \Comment{Minimize MSE}
        \State $\text{MSEvals}[k], \text{genExps}[k] \leftarrow \texttt{fitCoefficients} \left(\hat{\mathbf{e}}_k(\mathbf{x}), \mathcal{D} \right)$
    \EndFor
    \State $\tilde{f}(\mathbf{x}), \tilde{\mathbf{e}}(\mathbf{x}) \leftarrow \texttt{getTopExpr}(\text{genExps}, \text{MSEvals})$
    \State \Return $\tilde{f}(\mathbf{x})$, $\tilde{\mathbf{e}}(\mathbf{x})$
\EndFunction
\end{algorithmic}
\caption{MSSP applied to 1-D problems}
\label{alg:1DMSSP}
\end{algorithm}

\subsection{Experimental Results on 1-D Synthetic Problems}

\begin{figure}[t]
    \centering
    \includegraphics[width=.9\columnwidth]{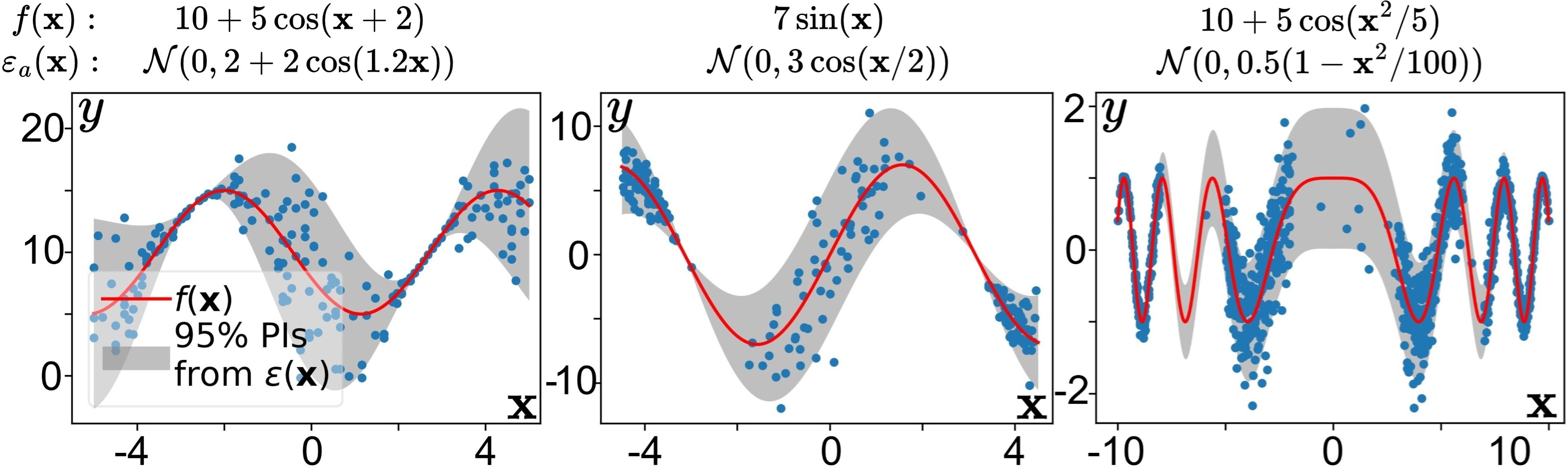}
    \vspace{-1ex}
    \caption{Initial \texttt{cos}, \texttt{hetero}, and \texttt{cosqr} datasets and the ideal 95\% PIs calculated from $\varepsilon_a(\mathbf{x})$ across the domain.}
    \vspace{-1ex}
    \label{fig:datasets}
\end{figure}

We considered three synthetic 1-D problems~\cite{adaptiveS}: \texttt{cos}, \texttt{hetero}, and \texttt{cosqr}.
All problems are affected by heteroscedastic noise (see Fig.~\ref{fig:datasets}).
For each case, we generated incomplete datasets as initial states to produce areas with low data density, which results in high epistemic uncertainty.

We evaluated the effectiveness of our SR approach within an AS framework called Adaptive Sampling with Prediction-Interval Neural Networks (ASPINN)~\cite{adaptiveS}.
ASPINN selects samples intelligently that contribute most to reducing uncertainty using prediction interval-generation NNs and Gaussian processes.
Hence, we analyze how the learned symbolic expressions evolve over AS iterations and assess whether they converge toward the underlying data-generating functions. 

At each AS iteration, $it$, we applied Algorithm~\ref{alg:1DMSSP} to recover an expression $\tilde{f}_{it}$ that approximates and explains the predictive function of the trained NN $\hat{f}_{it}$.
The approach constructs a collection of sets from $\mathcal{D}_{it} = (\mathbf{X}_{obs}^{(it)}, \mathbf{y}_{obs}^{(it)})$, each of which contains a random portion of the input domain.
It then recovers a skeleton that generalizes across the subdomain variations and from which a full mathematical expression $\tilde{f}_{it}$ is derived. 

To ensure consistency with the pre-trained Multi-Set Transformer~\cite{SeTGAP},
we used $N_S=10$ input sets or subdomains per MSSP instance, each containing $n=3000$ points.
Across all experiments, the Multi-Set Transformer was configured with a beam size $n_B=3$, generating $n_{\text{cand}}=5$ skeleton candidates per instance. 
Increasing these values was found to offer no additional benefit in terms of discovering distinct top-performing candidates.
Each skeleton was subsequently fitted to the observed dataset $\mathcal{D}_{it}$, and the final expression was selected based on the lowest mean squared error (MSE). 

Table~\ref{tab:SR-AS} summarizes the expressions recovered by SeTGAP at AS iterations $it \in \{1, 5, 10, \dots, 50\}$ for each tested problem.
At each iteration, the dataset $\mathbf{X}_{obs}^{(it)}$ is augmented with five samples using ASPINN, following the configuration detailed in \cite{adaptiveS}.
For each expression, we also report the predicted MSE on the entire dataset, computed using the learned expression $\tilde{f}_{it}$.
Highlighted cells indicate that the identified expression matches the functional form of the underlying function $f$. 

\begin{figure}[!t]
    \centering
    \includegraphics[width=.9\linewidth]{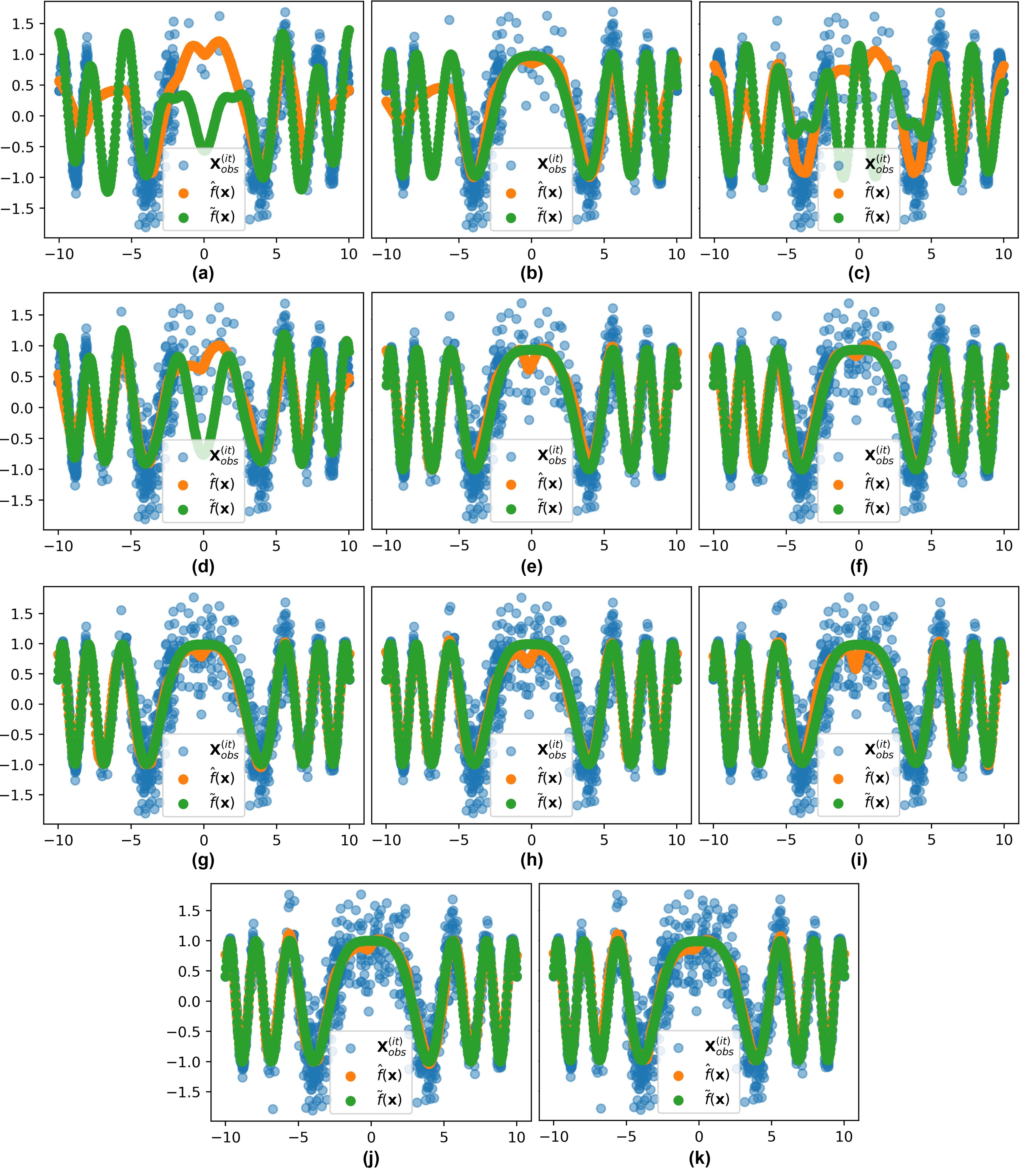}
    \vspace{-1ex}
    \caption{$\hat{f}_{it}(\mathbf{x})$ vs. $\tilde{f}_{it}(\mathbf{x})$ throughout the AS process for \texttt{cosqr}. \textbf{(a)} $it\!\!=\!\!1$. \textbf{(b)} $it\!\!=\!\!5$. \textbf{(c)} $it\!\!=\!\!10$. \textbf{(d)} $it\!\!=\!\!15$. \textbf{(e)} $it\!\!=\!\!20$. \textbf{(f)} $it\!\!=\!\!25$. \textbf{(g)} $it\!\!=\!\!30$. \textbf{(h)} $it\!\!=\!\!35$. \textbf{(i)} $it\!\!=\!\!40$. \textbf{(j)} $it\!\!=\!\!45$. \textbf{(k)} $it\!\!=\!\!50$.}
    \label{fig:SR-AScosqr}
    \vspace{-3ex}
\end{figure}

Finally, Fig.~\ref{fig:SR-AScosqr} compares the predicted curves obtained from the model $\hat{f}_{it}$ with those generated by the corresponding symbolic expressions $\tilde{f}_{it}$ at each iteration.
Early iterations often yield inaccurate or overfitted expressions due to sparse data coverage, but as AS progresses, the symbolic models stabilize and converge toward compact expressions that match the true function in both form and performance.

\begin{table}[t]
    \centering
    \small 
    \caption{Identified expressions during the AS process}
    \label{tab:SR-AS}
    \vspace{-1ex}
    \resizebox{1\columnwidth}{!}{
    \begin{tabular}{|c|cc|cc|cc|}
        \hline
        
        \multirow{2}{*}{\normalsize{$it$}} & \multicolumn{2}{c|}{\texttt{cos}} & \multicolumn{2}{c|}{\texttt{hetero}} & \multicolumn{2}{c|}{\texttt{cosqr}} \\ \cline{2-7} 
         & \multicolumn{1}{c|}{\normalsize{$\tilde{f}_{it}$}} & \normalsize{MSE} & \multicolumn{1}{c|}{\normalsize{$\tilde{f}_{it}$}} & \normalsize{MSE} & \multicolumn{1}{c|}{\normalsize{$\tilde{f}_{it}$}} & \normalsize{MSE} \\ \hline
        1 & 
        \multicolumn{1}{c|}{\begin{tabular}[c]{@{}c@{}}$5.153\sin(62.757\sqrt{1 - 0.032\mathbf{x}} $ \\ $+ 5.994) + 9.933$\end{tabular}} & 1.502 & \multicolumn{1}{c|}{\begin{tabular}[c]{@{}c@{}}$3.940\mathbf{x}\tanh(0.006\mathbf{x} - 6.212)$ \\ $ + 10.742\tanh(1.777\mathbf{x} - 0.123)$\end{tabular}} & 1.508  & \multicolumn{1}{c|}{\begin{tabular}[c]{@{}c@{}}$0.145 \mathbf{x} \sin(2.643 \mathbf{x} + 6.235) +$ \\ $ 0.564 \sin(1.732 \mathbf{x} + 10.919)$\end{tabular}} & 0.325 \\ \hline
        
        5 & \multicolumn{1}{c|}{\cellcolor{gray!30}\begin{tabular}[c]{@{}c@{}}$10.001 - 5.037\sin(1.002\mathbf{x} + $ \\ $12.989)$\end{tabular}} & 1.585 & \multicolumn{1}{c|}{\cellcolor{gray!30} \begin{tabular}[c]{@{}c@{}}$7.261\sin(0.998\mathbf{x} - 6.327) $ \\ $- 0.276$\end{tabular}} & 1.358 &  \multicolumn{1}{c|}{\cellcolor{gray!30}\begin{tabular}[c]{@{}c@{}}$0.012 - 0.983\cdot$ \\ $\sin(0.197\mathbf{x}^2 + 4.911)$\end{tabular}} & 0.222 \\ \hline
        
        10 & \multicolumn{1}{c|}{\begin{tabular}[c]{@{}c@{}}$\sqrt{1 - 0.971\sin(0.974\mathbf{x} + 6.55)} $ \\ $ \cdot8.973 + 2.123$\end{tabular}} & 1.768 & 
        \multicolumn{1}{c|}{ \cellcolor{gray!30}\begin{tabular}[c]{@{}c@{}}$7.203\sin(1.0\mathbf{x} + 18.809) $ \\ $- 0.261$\end{tabular}} & 1.300 &
        \multicolumn{1}{c|}{\begin{tabular}[c]{@{}c@{}}$-1.271\sin(1.201\mathbf{x} + 6.256)^2 $ \\ $- 0.495\sin(3.242\mathbf{x}$ \\ $ + 10.966) + 0.655$\end{tabular}} & 0.409 \\ \hline
        
        15 & \multicolumn{1}{c|}{\begin{tabular}[c]{@{}c@{}}$5.616\sin(0.508\mathbf{x} - 3.803)^2 +$ \\ $2.295\sin(0.973\mathbf{x} + 10.024) $ \\ $+ 7.281$\end{tabular}} & 1.811 & 
        \multicolumn{1}{c|}{\cellcolor{gray!30}\begin{tabular}[c]{@{}c@{}}$7.123\sin(\mathbf{x} - 18.942) $ \\ $- 0.025$\end{tabular}} & 1.256 & 
        \multicolumn{1}{c|}{\begin{tabular}[c]{@{}c@{}}$-0.016\mathbf{x}^2\sin(2.849\mathbf{x} +  1.598) + $ \\ $0.053 \mathbf{x}\sin(1.122 \mathbf{x} - 0.075)$ \\ $ + 0.762 \sin(1.745 \mathbf{x} + 4.757)$\end{tabular}} & 0.277 \\ \hline
        
        20 & \multicolumn{1}{c|}{\cellcolor{gray!30}\begin{tabular}[c]{@{}c@{}}$10.002 - 4.987\cdot$ \\ $\sin(0.995\mathbf{x} - 5.861)$\end{tabular}} & 1.866 & 
        \multicolumn{1}{c|}{\cellcolor{gray!30}\begin{tabular}[c]{@{}c@{}}$7.202\sin(0.999\mathbf{x} - 18.820) $ \\ $- 0.190$\end{tabular}} & 1.232 & 
        \multicolumn{1}{c|}{\cellcolor{gray!30}\begin{tabular}[c]{@{}c@{}}$0.993\sin(0.199\mathbf{x}^2 + 1.612) $ \\ $+ 0.003$\end{tabular}} & 0.220 \\ 
        \hline
        
        25 & \multicolumn{1}{c|}{\cellcolor{gray!30}\begin{tabular}[c]{@{}c@{}}$10.004 - 5.001\cdot$ \\ $\sin(0.996\mathbf{x} - 5.860)$\end{tabular}} & 1.967 & 
        \multicolumn{1}{c|}{\cellcolor{gray!30}\begin{tabular}[c]{@{}c@{}}$7.149\sin(0.999\mathbf{x} - 18.831) $ \\ $- 0.128$\end{tabular}} & 1.185 & 
        \multicolumn{1}{c|}{\cellcolor{gray!30}\begin{tabular}[c]{@{}c@{}}$0.994\sin(0.200\mathbf{x}^2 + 1.608) $ \\ $+ 0.004$\end{tabular}} & 0.222 \\ 
        \hline
        
        30 & \multicolumn{1}{c|}{\cellcolor{gray!30}\begin{tabular}[c]{@{}c@{}}$10.037 - 4.998\cdot$ \\ $\sin(0.996\mathbf{x} - 6.696)$\end{tabular}} & 1.988 & 
        \multicolumn{1}{c|}{\cellcolor{gray!30}\begin{tabular}[c]{@{}c@{}}$7.217\sin(0.999\mathbf{x} - 0.033)$ \\ $- 0.218$\end{tabular}} & 1.206 & 
        \multicolumn{1}{c|}{\cellcolor{gray!30}\begin{tabular}[c]{@{}c@{}}$0.991\sin(0.199\mathbf{x}^2 $ \\ $+ 1.623)$\end{tabular}} & 0.223\\ \hline
        
        35 & \multicolumn{1}{c|}{\cellcolor{gray!30}\begin{tabular}[c]{@{}c@{}}$9.987 + 5.014\cdot$ \\ $\sin(0.999\mathbf{x} + 16.138)$\end{tabular}} & 2.061 & 
        \multicolumn{1}{c|}{\cellcolor{gray!30}\begin{tabular}[c]{@{}c@{}}$7.138\sin(1.000\mathbf{x} + 18.835)$ \\ $- 0.093$\end{tabular}} & 1.223 & 
        \multicolumn{1}{c|} {\cellcolor{gray!30}\begin{tabular}[c]{@{}c@{}}$-0.997\sin(0.200\mathbf{x}^2 $ \\ $+ 17.315)$\end{tabular}} & 0.221\\ \hline
        
        40 & \multicolumn{1}{c|}{\cellcolor{gray!30}\begin{tabular}[c]{@{}c@{}}$10.002 + 5.003\cdot$ \\ $\sin(0.997\mathbf{x} - 15.285)$\end{tabular}} & 2.043 & 
        \multicolumn{1}{c|}{\cellcolor{gray!30}\begin{tabular}[c]{@{}c@{}}$7.128\sin(1.000\mathbf{x} + 18.838) $ \\ $- 0.071$\end{tabular}} & 1.216 & 
        \multicolumn{1}{c|}{\cellcolor{gray!30}\begin{tabular}[c]{@{}c@{}}$0.986\sin(0.198\mathbf{x}^2 $ \\ $+ 1.688) + 0.004$\end{tabular}} & 0.223\\ \hline
        
        45 & \multicolumn{1}{c|}{\cellcolor{gray!30}\begin{tabular}[c]{@{}c@{}}$10.011 + 5.043\cdot$ \\ $\sin(1.000\mathbf{x} - 9.003)$\end{tabular}} & 2.051 & 
        \multicolumn{1}{c|}{\cellcolor{gray!30}\begin{tabular}[c]{@{}c@{}}$- 7.097\sin(1.000\mathbf{x} -3.131) $ \\ $- 0.074$\end{tabular}} & 1.237 & 
        \multicolumn{1}{c|}{\cellcolor{gray!30}\begin{tabular}[c]{@{}c@{}}$1.0\sin(0.200\mathbf{x}^2 + 1.612) $ \\ $+ 0.002$\end{tabular}} & 0.228\\ \hline
        
        50 & \multicolumn{1}{c|}{\cellcolor{gray!30}\begin{tabular}[c]{@{}c@{}}$10.055 - 5.005\cdot$ \\ $\sin(0.999\mathbf{x} + 12.997)$\end{tabular}} & 2.055 & 
        \multicolumn{1}{c|}{\cellcolor{gray!30}\begin{tabular}[c]{@{}c@{}}$-7.150\sin(1.000\mathbf{x} + 15.689) $ \\ $- 0.135$\end{tabular}} & 1.244 & 
        \multicolumn{1}{c|}{\cellcolor{gray!30}\begin{tabular}[c]{@{}c@{}}$0.990\sin(0.199\mathbf{x}^2 - 17.217) $ \\ $+ 0.004$\end{tabular}} & 0.231\\ \hline
        \end{tabular}%
    }
    \vspace{-1.5ex}
\end{table}

\section{Winter Wheat N-response Curves} \label{sec:Nresponse}

Let $y_{(lat, lon)}$ represent the observed yield at a field site with coordinates $(lat, lon)$.
Furthermore, let $\mathbf{X}_{(lat, lon)}$ represent a set of multiple covariate factors that describe the state of the field at position $(lat, lon)$, and potentially its neighboring areas.
The underlying yield function of the field is denoted as $f(\cdot)$ and $y_{(lat, lon)} = f(\mathbf{X}_{(lat, lon)})$.
In practice, $f$ is a complex multivariate system with unknown functional form.
Nevertheless, tasks like N-rate optimization, which allows for profit maximization and environmental impact minimization~\cite{hegedus}, do not require estimating the full functional form of $f(\mathbf{X}_{(lat, lon)})$.
Instead, N-rate optimization only analyzes the functional relationship between the N-rate and the predicted yield values, typically represented using N-response curves.

This section aims to estimate the functional form of N-response curves for one winter wheat dryland field, referred to as Field A. 
Traditionally, N-response curves are modeled using a single parametric function for the entire field~\cite{bullock94,CF}. 
However, previous studies suggest that the functional form of these curves varies across different field regions due to factors such as terrain slope and soil composition~\cite{responsivity,MZ}.


\subsection{Parametric N-response Curve Learning}
\vspace{-.5ex}
Building on~\cite{MZ}, 
which clusters Field A into four MZs, we assume that all sites within a given MZ share the same functional form due to the shape similarity of the curves within each cluster. 
An N-response curve exhibits the estimated crop relative yield values $ry$ corresponding to a specific field site in response to all admissible fertilizer rates $\mathbf{x}^{Nr}$. 
For winter wheat, we impose a fixed N-rate range $\mathbf{x}^{Nr}$; i.e., between $0$ and $150$ pounds per acre (lbs/acre).
Consequently, estimating the functional form of N-response curves for an MZ can be formulated as an MSSP. 
Additional details, such as the early-yield prediction problem, its formulation using a 2D–3D convolutional NN regression model, the dataset, and the N-response curve generation process, can be found in~\cite{MZ}.

Given the pre-computed N-response curves for all field sites, $\mathbf{R}$, we generate skeleton expressions that capture the mathematical behavior of the N-response curves associated with the $z$-th MZ, denoted by $\mathbf{R}_z$.
The curves are standardized using $z$-score normalization, and the N rates are rescaled to the range $[-10, 10]$ to match the input format required by the Multi-Set Transformer.
The procedure is outlined in Algorithm~\ref{alg:Nresp_sk}.
In Line~\ref{alg:selectNcurves}, the function $\texttt{selectRandomCurves}(\mathbf{R}_z, N_s)$ is used to select $N_s$ N-response curves at random, forming the collection $\tilde{\mathbf{D}}_{\mathbf{R}_z}$. 
These selected curves are then input to a slightly modified version of Algorithm~\ref{alg:1DMSSP} to generate a candidate skeleton.
Note that Algorithm~\ref{alg:1DMSSP} constructs the collection $\mathbf{D}$ by sampling $N_S$ subsets from a single given curve.
Conversely, here, $N_S$ distinct curves are provided, from each of which a subset is sampled randomly, resulting in a collection of $N_S$ subsets derived from different curves but still sharing a common underlying structure.
This process is repeated $n_{\text{cand}}$ times, each with a newly sampled $\tilde{\mathbf{D}}_{\mathbf{R}_z}$, 
thereby enhancing diversity and promoting discovery of more generalizable skeletons. 
The resulting list of skeletons is denoted by $\texttt{genSks}_z =  \{ \hat{\mathbf{e}}_1(\mathbf{x}^{Nr}), \dots, \hat{\mathbf{e}}_{|\texttt{genSks}_z|}(\mathbf{x}^{Nr}) \}$. 
These skeletons are selected based on their ability to minimize the MSE after fitting a randomly selected test collection $\tilde{\mathbf{D}}_{\mathbf{R}_z}^{(\text{test})}$. 

\setlength{\textfloatsep}{8pt}
\begin{algorithm} [!t]
\scriptsize
\renewcommand{\algorithmicrequire}{\textbf{Input:}}  
\renewcommand{\algorithmicensure}{\textbf{Output:}} 
\begin{algorithmic}[1]
    \Require Aligned N-response curves of the $z$-th MZ, $\mathbf{R}_z$; number of input sets $N_S$; \hspace*{1.6em}number of skeleton candidates $n_{\text{cand}}$
    \Ensure Generated skeleton for the $z$-th MZ
    \vspace{-1.5ex}
    \Statex \makebox[0pt][l]{\hspace*{-\algorithmicindent}\rule{\dimexpr\linewidth+\algorithmicindent}{0.4pt}}
\Function{generateNrespSks}{$\mathbf{R}_z, N_s, n_{\text{cand}}$}
    \State $\text{genSks}_z \leftarrow [\,]$ 
    \For { each $i \in (1, n_{\text{cand}})$}
        \State $\tilde{\mathbf{D}}_{\mathbf{R}_z} \leftarrow \texttt{selectRandomCurves}(\mathbf{R}_z, N_s)$ \label{alg:selectNcurves}
        \State $ \text{genSks}_z.\texttt{append}(\texttt{Generate1DExpr}_{\text{modified}}(\tilde{\mathbf{D}}_{\mathbf{R}_z}))$
    \EndFor
    \State $\text{genSks}_z \leftarrow \text{removeDuplicates}(\text{genSks}_z)$  
    \State $\tilde{\mathbf{D}}_{\mathbf{R}_z}^{(\text{test})} \leftarrow \texttt{selectRandomCurves}(\mathbf{R}_z, N_s)$
    \State $\text{MSEvals}_z \leftarrow \text{zeros}(|\text{genSks}_z|)$
    \For { each $k \in (1, n_{\text{cand}})$} 
        \State $\text{MSEvals}_z[k] \leftarrow \texttt{fitCoefficients} (\hat{\mathbf{e}}_k(\mathbf{x}^{Nr}), \tilde{\mathbf{D}}_{\mathbf{R}_z}^{(\text{test})})$
    \EndFor
    \State \Return $\texttt{getTopSk}(\text{genSks}_z, \text{MSEvals})$
\EndFunction
\end{algorithmic}
\caption{N-response Curve Skeleton Generation}
\label{alg:Nresp_sk}
\end{algorithm}

\begin{figure}[!t]
    \centering
    \includegraphics[width=.9\columnwidth]{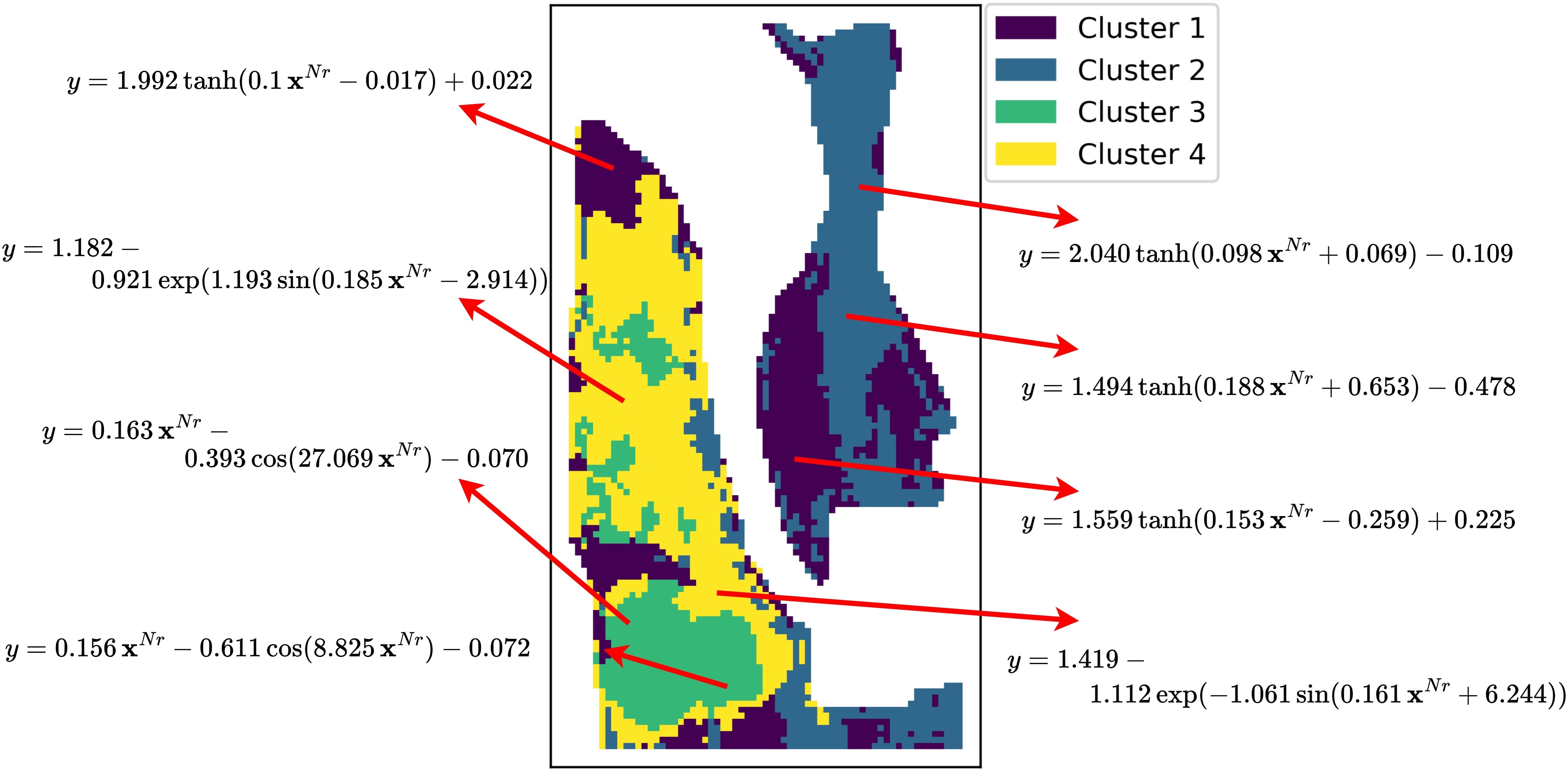}
    \vspace{-1ex}
    \caption{Fitted N-response curves using the identified skeletons.} 
    \label{fig:rcurves_skeletons}
    \vspace{-3ex}
\end{figure}

\begin{figure}[!t]
    \centering
    \includegraphics[width=\columnwidth]{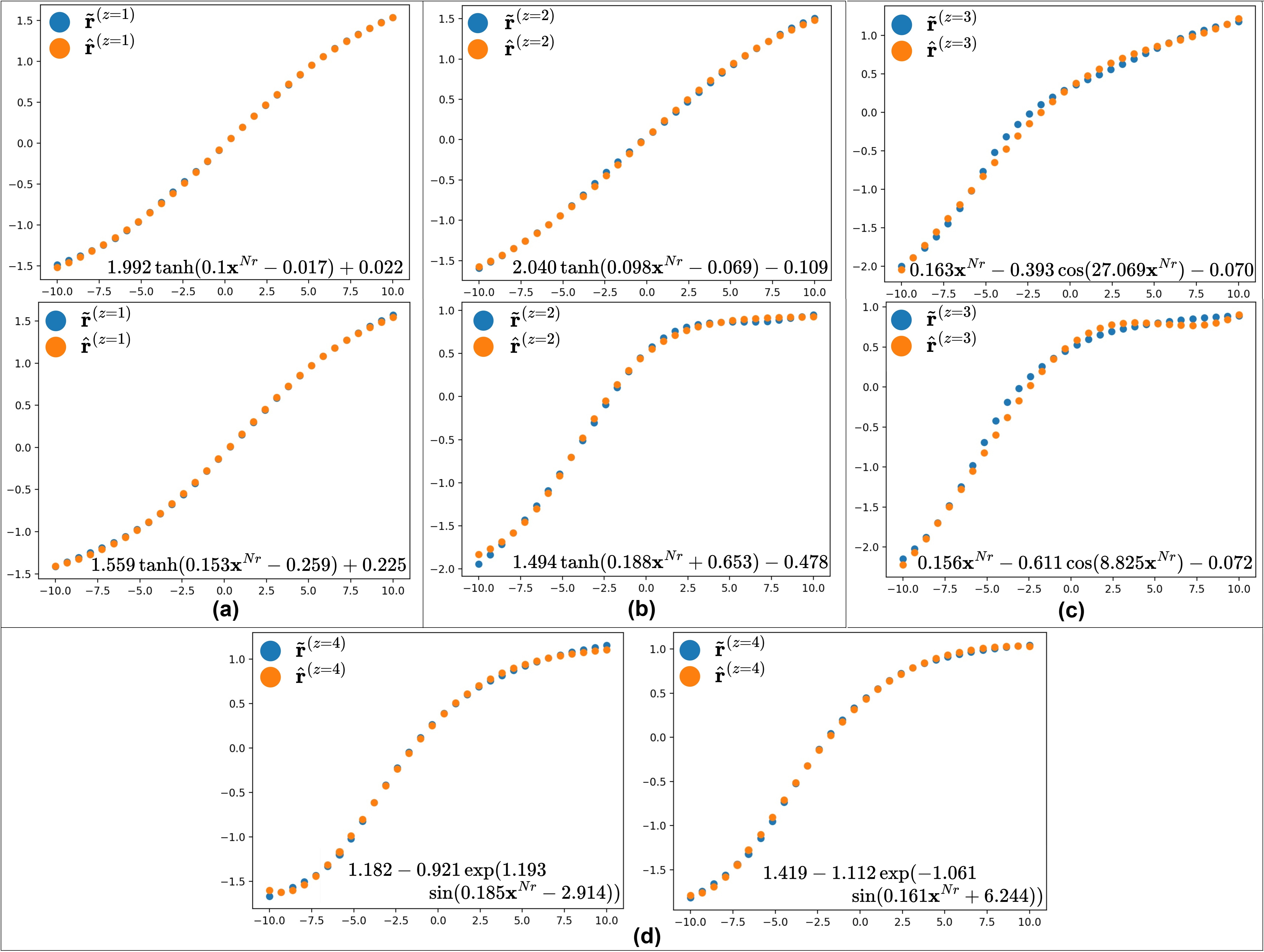}
    \vspace{-3ex}
    \caption{Comparison between NN-generated N-response curves $\tilde{\mathbf{r}}$ and fitted curves $\hat{\mathbf{r}}$ from \textbf{(a)} MZ 1, \textbf{(b)} MZ 2, \textbf{(c)} MZ 3, and \textbf{(d)} MZ 4. Equation $\hat{\mathbf{r}}$ at the bottom of each plot.} 
    \label{fig:rcurves_comparison}
    \vspace{-1ex}
\end{figure}

For our experiments, we set $n_{\text{cand}} \!\!=\!\! 5$, as higher values did not yield more distinct skeletons.
Table~\ref{tab:results_Nresponse} shows the skeletons derived for each MZ.
Furthermore, we evaluated the suitability of the obtained skeletons for each MZ and compared them to two traditional N-response models: quadratic-plateau~\cite{bullock94} and exponential~\cite{CF}.
For each method, field site, and MZ, we fitted the skeleton's coefficient values to minimize the distance to the corresponding N-response curve.
We used the GA-based optimization function $\hat{\mathbf{r}}_j^{(z)} \!\!=\!\! \texttt{fitCoefficients}(\hat{\mathbf{e}}(\mathbf{x}^{Nr}), \tilde{\mathbf{r}}_j^{(z)})$ that returns the fitted function $\hat{\mathbf{r}}_j^{(z)}$ that minimizes the MSE error with respect to the $j$-th aligned curve in $\mathbf{R}$, $\tilde{\mathbf{r}}_j^{(z)}$.  
We report the mean error $\bar{r}$ obtained considering all sites within each MZ: $\bar{r} = \frac{1}{|\mathbf{R}_z|} \sum_{\tilde{\mathbf{r}}_j^{(z)} \in \mathbf{R}_z} | \hat{\mathbf{r}}_j^{(z)} -  \tilde{\mathbf{r}}_j^{(z)} |.$
Fig.~\ref{fig:rcurves_skeletons} shows two fitted curves for each MZ, derived from the skeletons identified by our approach. 
Fig.~\ref{fig:rcurves_comparison} compares these fitted curves against the corresponding aligned N-response curves.

\begin{table}[t]
\vspace{-1ex}
    \caption{Skeleton prediction results comparison for Field A}
    \label{tab:results_Nresponse}
    \vspace{-1ex}
\centering
\resizebox{.7\columnwidth}{!}{%
\begin{tabular}{|c|c|c|c|}
\hline
{Method} & {MZ} & {Functional Form} & {$\bar{r}$} \\ \Xhline{3\arrayrulewidth}
Quadratic-plateau & \multirow{3}{*}{1} & 
$c_1 + c_2\,(\min(\mathbf{x}^{Nr}, c_3) + c_3)^2$ & 0.0695 \\ \cline{1-1} \cline{3-4}
Exponential & & $c_1 (1-\exp(c_2 + c_3\, \mathbf{x}^{Nr})) + c_4$ & 0.2303 \\ \cline{1-1} \cline{3-4}
\textbf{SeTGAP} & & $c_1 + c_2 \, \tanh(c_3 + c_4\, \mathbf{x}^{Nr})$ & \textbf{0.0620} \\ \Xhline{3\arrayrulewidth}
Quadratic-plateau & \multirow{3}{*}{2} & 
$c_1 + c_2\,(\min(\mathbf{x}^{Nr}, c_3) + c_3)^2$ & 0.0725 \\ \cline{1-1} \cline{3-4}
Exponential & & $c_1 (1-\exp(c_2 + c_3\, \mathbf{x}^{Nr})) + c_4$ & 0.1825 \\ \cline{1-1} \cline{3-4}
\textbf{SeTGAP} & & $c_1 + c_2\, \tanh(c_3 + c_4\, \mathbf{x}^{Nr})$ & \textbf{0.0355} \\ \Xhline{3\arrayrulewidth}
Quadratic-plateau & \multirow{3}{*}{3} & 
$c_1 + c_2\,(\min(\mathbf{x}^{Nr}, c_3) + c_3)^2$ & 0.1028 \\ \cline{1-1} \cline{3-4}
Exponential & & $c_1 (1-\exp(c_2 + c_3\, \mathbf{x}^{Nr})) + c_4$ & 0.1965 \\ \cline{1-1} \cline{3-4}
\textbf{SeTGAP} & & $c_1 + c_2\,\mathbf{x}^{Nr} + c_3\cos(c_4 + c_5\,\mathbf{x}^{Nr})$ & \textbf{0.0683} \\ \Xhline{3\arrayrulewidth}
Quadratic-plateau & \multirow{3}{*}{4} & 
$c_1 + c_2\,(\min(\mathbf{x}^{Nr}, c_3) + c_3)^2$ & 0.0615 \\ \cline{1-1} \cline{3-4}
Exponential & & $c_1 (1-\exp(c_2 + c_3\, \mathbf{x}^{Nr})) + c_4$ & 0.2249 \\ \cline{1-1} \cline{3-4}
\textbf{SeTGAP} & & $c_1 + c_2\exp(c_3\sin(c_4 + c_5\,\mathbf{x}^{Nr}))$ & \textbf{0.0448} \\ \hline
\end{tabular}%
}
\vspace{-2ex}
\end{table}

\section{Discussion} \label{discussion}

Our work focuses on learning parametric expressions that describe N-response curves through a multi-set SR strategy. 
We first introduced this strategy in a controlled setting to address 1-D expression discovery under epistemic uncertainty. 
To systematically evaluate its behavior across different uncertainty levels and track its evolution, we relied on synthetic functions within an AS framework, since such controlled analysis is not feasible with real agricultural data. 
This setup allowed us to isolate the effects of data sparsity and uncertainty on symbolic recovery, providing a clear basis for understanding the strengths and limitations of the proposed approach.

The results presented in Table~\ref{tab:SR-AS} demonstrate that the estimated expressions began consistently matching the true functional form of the target function by iteration $it\!\!=\!\!20$ across all tested cases. 
Although the correct symbolic structure was occasionally identified at earlier stages (e.g., at $it\!\!=\!\!5$), these discoveries tended to be unstable, as subsequent iterations often produced alternative expressions. 
This instability highlights the inherent challenge of model identification when domain coverage is limited or predictive uncertainty remains high.

In the early iterations of the AS process, the observed data are sparse and unevenly distributed across the input domain. 
For instance, Fig.\ref{fig:SR-AScosqr}.b illustrates the prediction generated by $\hat{f}_{it}$ at $it\!\!=\!\!5$, where limited observations in the central region of the domain result in a relatively smooth estimated response by the model $\hat{f}_{it}$.
In such cases, we can occasionally recover a correct or near-correct functional form because the prediction lacks fine-grained variations that would otherwise complicate the model discovery task. 
However, as sampling proceeds and additional points are collected in previously underrepresented regions, local variations in the curve predicted by $\hat{f}_{it}$ begin to emerge, as shown in Fig.\ref{fig:SR-AScosqr}.c for $it\!\!=\!\!10$. 
These local fluctuations can mislead the SR process, causing the symbolic models to temporarily favor more complex or distorted expressions that attempt to capture these finer structures.

As the AS process strategically targets regions of high epistemic uncertainty, the overall domain coverage improves, progressively eliminating major gaps and inconsistencies in the input space. 
Consequently, our SeTGAP variant gains access to increasingly informative and representative subsets of the domain, allowing it to generate symbolic models that are not only more accurate but also more stable over time. 


In certain cases, we uncovered the correct underlying functional form even when the prediction model $\hat{f}_{it}$ exhibited notable inaccuracies.
Continuing the analysis of Fig.~\ref{fig:SR-AScosqr}.b, we observe that the prediction produced by $\hat{f}_{it=5}$ deviates significantly from the expected behavior in the region $\mathbf{x} \in [-7, -5]$, primarily due to sparse data coverage in that interval. 
Despite this local error, Algorithm~\ref{alg:1DMSSP} was able to identify the correct symbolic skeleton and, after fitting it to the available observations, produced a response closely resembling the true function.  
This robustness arises from our multi-set strategy, which involves randomly sampling diverse subregions across the domain under the assumption that they share a common functional form.
In this example, the majority of sampled subregions exhibited a strong $\sin(\mathbf{x}^2)$ behavior, allowing the symbolic regression process to prioritize consistent patterns over localized uncertainties and diminishing the impact of poorly sampled regions during skeleton discovery.

Having verified the effectiveness of the proposed multi-set strategy, we applied it to distill parametric expressions from NN-generated N-response curves across four delineated MZs. 
In this setting, the objective shifts from recovering known functions to extracting shared functional structures that characterize groups of curves with similar agronomic behavior.
Unlike traditional approaches that assume a single form for the entire field, modeling each MZ with a distinct functional form is agronomically justified, as different regions often exhibit varying soil types and terrain characteristics. 
This hypothesis is supported by the results in Table~\ref{tab:results_Nresponse}, which show that using tailored skeletons leads to lower fitting errors, indicating their greater suitability for modeling the field’s N-response curves.
From Fig.~\ref{fig:rcurves_comparison}, the close agreement in shape between the reference and fitted N-response curves confirms that the identified functional forms were appropriate for all cases.

The insight gained from the discovered models lies in revealing how the functional relationship between N inputs and yield response shifts across different MZs of the same field. 
By analyzing the differences in the equations generated for each MZ, agronomists gain a direct way to analyze spatial variations in N responsivity. 
For instance, discovering a steeper slope or a lower saturation plateau in a specific zone provides insight into that soil's carrying capacity and biological limits.

The proposed SR application advances PA by offering a data-driven yet interpretable alternative to opaque models, capturing complex local variations in N-response without predefining functional forms. 
As such, we applied our SeTGAP variant to model N-response curves at the MZ level.
These MZs were derived based on shape dissimilarities in fertilizer responsivity rather than yield productivity. 
These MZ-specific expressions captured diverse crop responses across spatial subregions, offering interpretable and data-driven alternatives to traditional parametric models. 
The resulting equations not only reflect localized agronomic dynamics but also serve as functional foundations for further analysis and simulation.

\section{Conclusion} \label{conclusion}

Modeling crop response to N fertilization is a central problem in PA, with direct implications for both economic efficiency and environmental sustainability. 
However, existing approaches either rely on rigid predefined functions or opaque models that limit interpretability, making it difficult to capture and understand site-specific response patterns.

In this work, we introduced a neuro SR approach to learn parametric N-response curves without assuming a predefined form. 
We proposed a multi-set strategy to recover shared symbolic structures under epistemic uncertainty and demonstrated its robustness through experiments on synthetic 1-D problems. 
We applied this approach to winter wheat data, extracting interpretable, MZ-specific expressions from NN-generated N-response curves. 
The results showed that the learned expressions closely match the original curves and improved fitting accuracy over traditional models.

This work opens several directions for future research. 
First, the proposed SR strategy constitutes an \textit{ad hoc} solution tailored to the specific challenges of this application, rather than a novel state-of-the-art SR methodology. 
Future work will focus on generalizing and formalizing this strategy, extending it to higher-dimensional settings and conducting comprehensive comparisons against existing SR approaches. 
In addition, while the results obtained for Field A demonstrate the practical potential of the method, they primarily serve as a proof of concept. 
A natural next step is to evaluate the approach across multiple fields, farms, and crop types to assess its robustness and generalizability under diverse agronomic conditions.

\vspace{-.5ex}
\bibliographystyle{IEEEtran}
\bibliography{references}

\end{document}